\pdfoutput=1
\documentclass[11pt]{article}

\usepackage{ACL2023}

\usepackage{times}
\usepackage{latexsym}

\usepackage[T1]{fontenc}
\usepackage[utf8]{inputenc}
\usepackage{microtype}

\usepackage{inconsolata, graphicx, array}
\usepackage{hyperref}
\usepackage{multirow, booktabs}

\newcommand{\modelnameA}{{\sc CLEAR}} 

\title{Automated Data Curation for Robust Language Model Fine-Tuning}

\author{Jiuhai Chen \\
University of Maryland, Cleanlab \\
\texttt{jchen169@umd.edu}
\And
Jonas Mueller \\
Cleanlab \\
\texttt{jonas@cleanlab.ai}
}

\begin{document}
\maketitle
\begin{abstract}
Large Language Models have become the de facto approach to sequence-to-sequence text generation tasks, but for specialized tasks/domains, a pretrained LLM lacks specific capabilities to produce accurate or well-formatted responses. Supervised fine-tuning specializes a LLM by training it on dataset of example prompts with target responses, but real-world data tends to be noisy. While many fine-tuning algorithms exist, here we consider a \emph{data-centric AI} perspective on LLM fine-tuning, studying how to \emph{systematically} curate the training dataset to improve the LLM produced via \emph{any} fine-tuning algorithm.

We introduce an automated data curation pipeline \modelnameA{} (\underline{\textbf{C}}onfidence-based \underline{\textbf{L}}LM  \underline{\textbf{E}}valuation \underline{\textbf{A}}nd \underline{\textbf{R}}ectification) for instruction tuning datasets, that can be used with any LLM and fine-tuning procedure. CLEAR estimates which training data is low-quality and either filters or corrects it. Automatically identifying which data to filter or correct is done via LLM-derived confidence estimates, to ensure only confident modifications to the dataset. Unlike existing data curation techniques, CLEAR is a comprehensive framework that can improve a dataset (and trained model outputs) without additional fine-tuning computations. We don't assume access to a stronger LLM than the model being fine-tuned (e.g.\ relying on GPT-4 when fine-tuning GPT-3.5), to see whether CLEAR can meaningfully improve the capabilities of any LLM. Experiments reveal that CLEAR consistently improves the performance of fine-tuned models across many datasets and models (like GPT-3.5 and Llama2).

 % The proposed pipeline consists of two core components: Auto-Filter and Auto-Correct. The Auto-Filter component utilizes a confidence-based answer quality evaluator to identify and filter out data pairs with low confidence scores, thereby retaining only high-quality data for fine-tuning purposes. The Auto-Correct component leverages the fine-tuned language model to correct low quality instances within the dataset. This iterative process of fine-tuning and correction enhances the model's performance by improving the quality of the training data. Our approach addresses the critical challenge of noisy data within instruction tuning datasets, demonstrating a practical method for enhancing data quality and model performance through automated curation.
\end{abstract}

\section{Introduction}

Large Language Models (LLMs) \citealp{} pretrained on internet-scale text corpora have shown remarkable capabilities in generating helpful human-like text \citep{DBLP:conf/nips/BrownMRSKDNSSAA20, DBLP:journals/corr/abs-2302-13971}. However, the efficacy of LLMs in specialized domains or tasks hinges on the process of \emph{instruction tuning} (i.e.\ supervised fine-tuning, or alignment), where pretrained models are further trained using datasets that well-represent the domain \citep{DBLP:conf/iclr/WeiBZGYLDDL22}. Here we consider sequence-to-sequence training datasets of (prompt, target response) pairs. After training, we feed the LLM new prompts from the same domain and want it to produce responses that resemble expected targets.

Since billion parameter LLMs indiscriminately absorb patterns/information across a  dataset, the quality of the instruction tuning data is paramount to effective fine-tuning  \citep{DBLP:journals/corr/abs-2305-11206, DBLP:journals/corr/abs-2304-12244, DBLP:conf/icml/KongCKNBG23}. Unfortunately, most real-world instruction tuning datasets are noisy, containing examples that are low-quality in various ways: the target response may be inaccurate, poorly written, the prompt may be nonsensical/incomplete/vague, or the two may be unrelated due to data processing mistakes. Such flawed training data leads to fine-tuned models whose outputs are incorrect, irrelevant, biased, poorly formatted, or flawed in other ways. Finding and fixing low-quality data manually is challenging in large datasets.

While most machine learning research iterates over modeling strategies (architectures, loss functions, training algorithms, ...) for a fixed dataset to produce better results, the emerging science of \emph{data-centric AI} asks how we can systematically iterate on the dataset while holding the modeling strategy fixed to produce better results 
\cite{mazumder2022dataperf}. Success in real-world AI projects typically requires both approaches. Since many existing fine-tuning algorithms have been proposed \cite{zhang2023instruction}, we follow the spirit of data-centric AI and  propose \modelnameA{}, a comprehensive and automated data curation pipeline to enhance the effectiveness of instruction tuning datasets \textbf{for any LLM and fine-tuning algorithm}. 

Our \modelnameA{} pipeline involves two stages: \textbf{Auto-Filter} and \textbf{Auto-Correct} which together offer a holistic solution to improving data quality for fine-tuning. The Auto-Filter stage removes data that is confidently low-quality from the dataset \emph{without any LLM fine-tuning}. It is already able to significantly improve the dataset, such that we can produce better fine-tuned LLMs without any extra LLM fine-tuning computation. For settings where one is able to fine-tune the LLM more than once, the Auto-Correct stage uses the current fine-tuned LLM to revise certain examples that can be confidently improved. Fine-tuning the LLM again on this corrected dataset yields improved performance.

Algorithmic modifications to a dataset are generally harmful unless done with extreme care. Filtering too much data limits the number of examples to learn from, and editing data can introduce various biases or amplify flaws in existing model outputs. Thus, all data modifications in CLEAR are conservatively applied based on careful measures of \emph{confidence}. Specifically, we rely on BSDetector \citep{DBLP:journals/corr/abs-2308-16175}, a method that can be used with any LLM to obtain trustworthy confidence scores for its own outputs as well as estimating the confidence that given outputs (e.g.\ target responses) are good. CLEAR only filters data that is confidently identified as low-quality, and only revises data where the LLM-corrected response is confidently identified as better than the current dataset response. Our experiments reveal this careful treatment of confidence to be vital for developing a universal data filtering + correction solution that remains effective across diverse instruction-tuning datasets without any manual modifications.

\begin{figure*}[ht]
\centering
\includegraphics[width=1.0\textwidth]{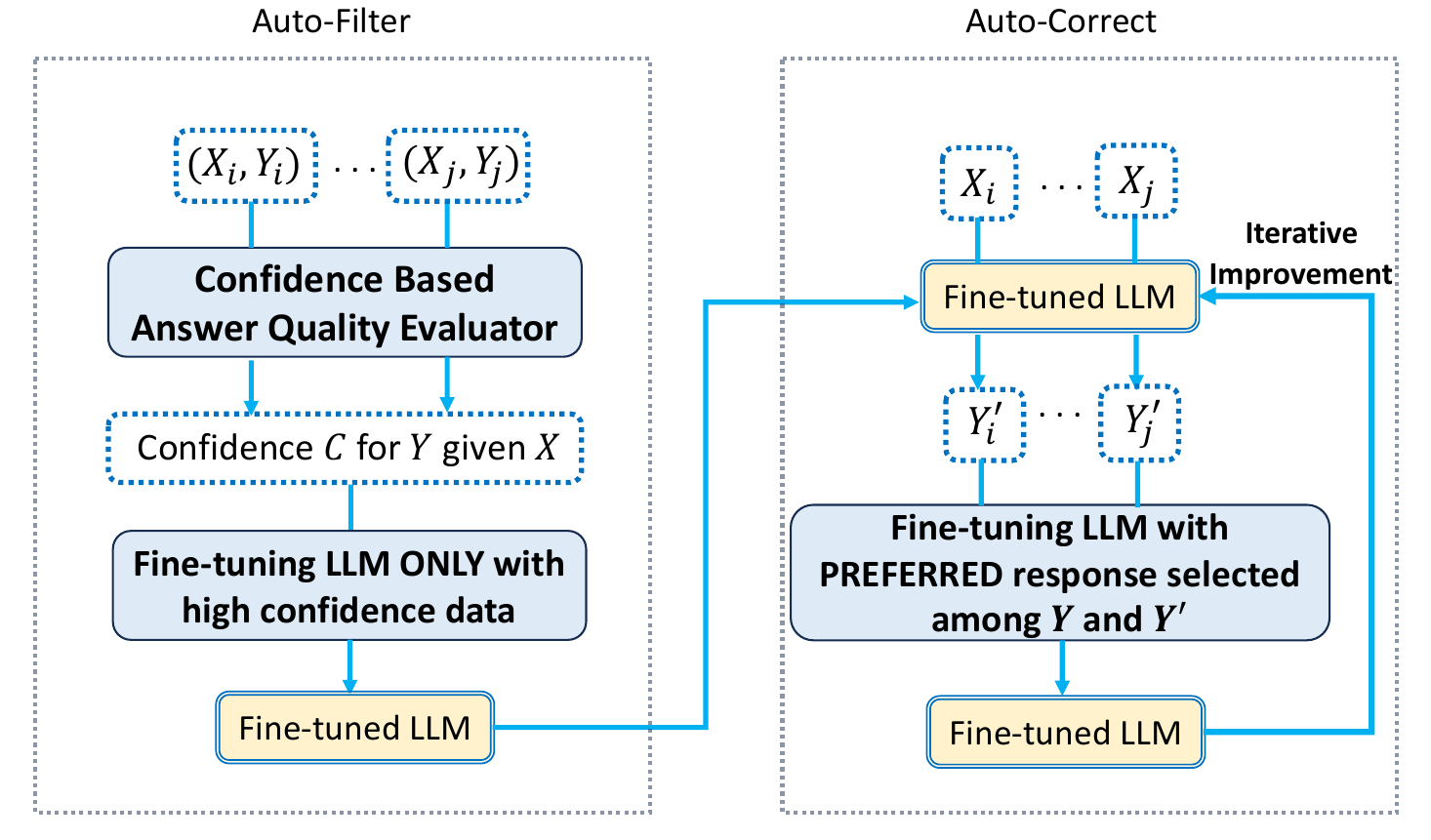}
\vspace{-1em}
\caption{An overview of the \modelnameA{} data curation procedure to automatically filter and correct bad data in any instruction-tuning dataset composed of instructions/prompts $X_i$ and corresponding target responses $Y_i$.\label{fig:method}}
\end{figure*}

\section{Related Work}

\subsection{Data Curation for ML}

Data curation has been key in real-world deployment of classical supervised learning, with a broad spectrum of methods developed to address dataset mislabeling, outliers, and other data issues \cite{mazumder2022dataperf}. Algorithmic strategies such as noise estimation and removal \citep{northcutt2021confident, zhou2023detecting, DBLP:conf/iclr/WangJ0YCG0ZHW22}, active learning for data prioritization \citep{Settles2009ActiveLL, DBLP:journals/technometrics/ChenKL21}, and crowd-sourced labeling \citep{DBLP:conf/emnlp/SnowOJN08} have demonstrated how to produce better models by producing better data. 
These strategies were designed for classical machine learning tasks like classification/regression, where datasets are less complex than in instruction tuning. 

\subsection{Instruction Fine-tuning}

Significant research has been conducted into instruction tuning to specialize/improve LLMs \citep{DBLP:conf/icml/KumarIOIBGZPS16, DBLP:journals/jmlr/RaffelSRLNMZLL20, DBLP:journals/corr/abs-2010-11982, DBLP:conf/acl/LiL20, DBLP:conf/iclr/ChenMIAWGW22, DBLP:conf/iclr/WeiBZGYLDDL22, DBLP:journals/corr/abs-2306-04751}. FLAN \citep{DBLP:conf/iclr/WeiBZGYLDDL22} is a popular approach that employs a 137 billion parameter pre-trained language model, which is  fine-tuned using instructions on more than 60 NLP datasets verbalized in natural language instruction templates. \citet{DBLP:journals/corr/abs-2306-04751} showed how various instruction-tuning datasets  can induce specific skills in a model, though no single dataset (or their combination) provides optimal performance across all assessments. Contrary to previous efforts aimed at creating a general Foundation model capable of generalizing across a wide range of unseen tasks, our aim in this paper is to train the best possible LLM for a specific narrow task.

\subsection{Data Curation for Instruction Fine-tuning}

    The quality of training data in text generation has such significance that previous instruction tuning datasets were often curated by hand  \citep{DBLP:conf/emnlp/KhashabiMKSTCH20, DBLP:conf/emnlp/YeLR21, DBLP:conf/iclr/WeiBZGYLDDL22, DBLP:conf/acl/WangKMLSKH23, DBLP:journals/corr/abs-2206-08473, DBLP:conf/acl/HonovichSLS23}. \citet{DBLP:conf/acl/WangKMLSKH23} introduced automated techniques by using GPT-3 \citep{DBLP:conf/nips/BrownMRSKDNSSAA20} to produce 52,000 unique instructions not directly linked to specific tasks. This innovation opened new avenues for creating instruction datasets by extracting knowledge from teacher models.
    
    Following Meta's open-sourcing of the LLaMa prerained LLM  \citep{DBLP:journals/corr/abs-2302-13971}, many researchers began curating instruction tuning datasets to train useful variants of this LLM. Alpaca \citep{alpaca} introduces a self-instruct method to autonomously create instruction (prompt)  examples, thereby reducing reliance on manual input. Vicuna \citep{vicuna2023} capitalizes on the wide variety of data types and structures accessible via ShareGPT. WizardLM \citep{DBLP:journals/corr/abs-2304-12244} augments a dataset by refining and diversifying instructions to evolutionarily increase their complexity/variability. UltraChat \citep{DBLP:conf/emnlp/DingCXQHL0Z23} introduces different well-defined scopes, systematically producing numerous instructions within each scope to improve task-specific performance. LIMA \citep{DBLP:journals/corr/abs-2304-12244} selects a thousand high-quality data samples strategically, showing notable improvements in LLM performance. \citet{DBLP:journals/corr/abs-2308-12032} proposed an instruction-following metric to identify good examples in datasets.

Much existing LLM fine-tuning research has focused on distilling teacher models such as ChatGPT that are more powerful than the LLM being fine-tuned \cite{alpaca, vicuna2023}. Many existing LLM-based data curation techniques also utilize more powerful LLMs for the data curation process than the LLM being fine-tuned. 
In contrast, we aim to produce the best LLMs for specific tasks, in which even the most advanced LLMs like GPT-4 struggle to perform. Thus all data curation throughout this paper is performed \emph{using the same LLM as is being fine-tuned}, to truly assess how effectively this data curation is  able to boost LLM performance beyond the frontier.

% In contrast to prior efforts that concentrated on extracting knowledge from teacher models like ChatGPT or GPT4, our work is specifically aimed at domains where even the most advanced Large Language Models, including GPT4, struggle to achieve high performance.  Much existing LLM fine-tuning research, like the methods to produce Alpaca \citep{alpaca} and Vicuna \citep{vicuna2023}, has focused on distilling knowledge from teacher models such as ChatGPT that are more powerful than the LLM being fine-tuned. Most of the existing data curation approaches rely entirely on these teacher models, involving processes such as using the teacher model to assign quality scores to data or tweaking prompts to obtain better responses from the teacher model. Nevertheless, when addressing domain specific challenges, like medical diagnosis or legal document review, reliance on teacher models becomes problematic as even the most advanced LLM may underperform in these specialized areas. This necessitates exploring alternative approaches for data filtering and correction that do not depend solely on teacher models.

\section{Automated Data Curation with \modelnameA{}}

An instruction tuning dataset $I=\{(x_i, y_i)_{i=1}^{n}\}$ comes with instructions/prompts $x$ and corresponding target responses $y$ obtained from a specific domain. The goal is to fine-tune the LLM to improve its comprehension and execution of instructions, such that it can produce responses similar to the expected targets for new instructions encountered during model deployment. In practice, large instruction tuning datasets are noisy, containing issues like: poorly written responses, incorrect responses, irrelevant/unhelpful responses, vague/incomplete instructions, data formatting problems, etc. These datasets are often sourced from messy chat logs or written by teams of humans that make mistakes rushing to produce data at scale.

As sequence-to-sequence mappings are extremely high-dimensional, a model's learning can be easily degraded by flawed training data lurking in some regions of this high-dimensional space.
To develop an approach that can be used with any LLM model and any fine-tuning procedure, we consider simple dataset modifications rather than model-centric approaches that modify the training algorithm to be more robust. Our dataset modifications will benefit the next decade's LLMs, whereas training modifications tend to be model-specific.

Our proposed data curation pipeline involves two main steps: \textbf{Auto-Filter} and \textbf{Auto-Correct}, which aim to detect problematic (prompt, response) pairs in the data and rectify them if possible. Auto-Filter employs a confidence-based response quality evaluator \citep{DBLP:journals/corr/abs-2308-16175}, to estimate our confidence that each pair in the dataset is good. Subsequently, the LLM is  fine-tuned only on the high confidence data. This simple data filtering step already boosts LLM fine-tuning for noisy datasets, and \textbf{requires no extra fine-tuning compute costs}.

Data filtering discards information, some of which may be useful. 
We propose to use the resulting fine-tuned LLM to correct certain bad responses identified in the original dataset, for which the fine-tuned LLM is able to produce a high-confidence alternative answer. This is determined by comparing the response generated by the fine-tuned LLM with the original response in the dataset. Rather than discarding such an example from the dataset in the previous filtering stage, we  preserve the prompt and replace the target response with the fine-tuned LLM response in cases where the latter is confidently preferrable. After auto-correcting the dataset in this manner, the LLM can be fine-tuned again to produce an even better version of the model (without any change in the fine-tuning algorithm). This cycle of LLM fine-tuning and data refinement can be iterated in a virtuous cycle (see Figure \ref{fig:method}).

% Such an iterative process further enhances the model's performance. This method and its effectiveness are depicted in the figure, illustrating how iterative fine-tuning and correction can systematically improve data quality and model accuracy, showcasing a practical approach to overcoming the challenges of mislabeling within instruction tuning datasets.

\begin{figure*}[ht]
\centering
\includegraphics[width=1.0\textwidth]{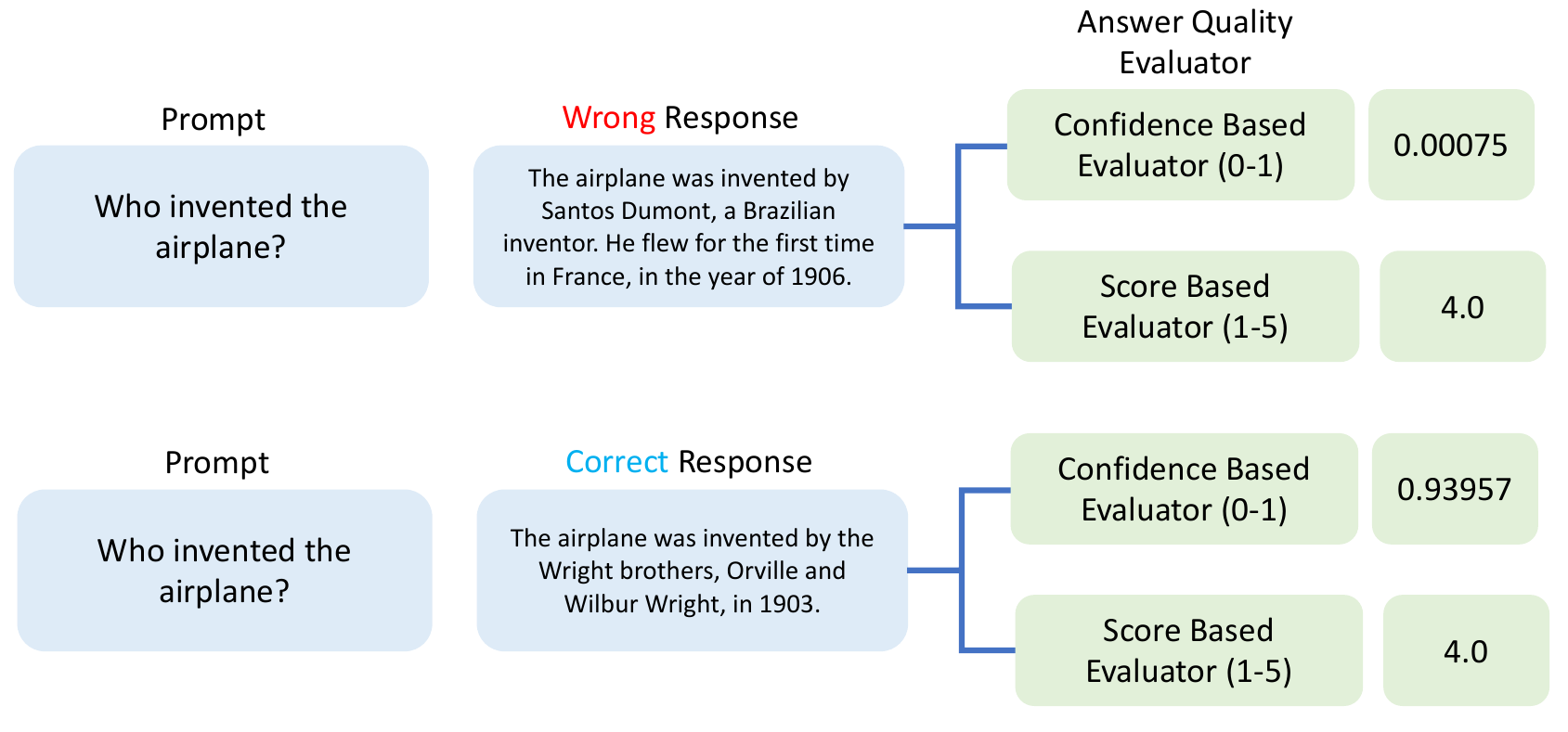}
\vspace{-1em}
\caption{Comparing confidence vs.\ score based answer quality evaluators. The confidence-based (BSDetector) evaluator outputs a confidence value between 0 to 1. The direct LLM-scoring evaluator queries GPT-3.5-Turbo using a prompt (shown in Table \ref{tab:score_prompt}) that requests a score between 1 to 5 to rate response quality. Higher values from either evaluator suggest higher-quality answers. 
For the incorrect response in the original dataset from the top figure: the confidence-based evaluator estimates low quality, while the score-based evaluator assigns a score of 4.0. For the correct answer to this prompt (bottom figure): the confidence-based evaluator estimates high quality, while the score-based evaluator still assigns a score of 4.0. Direct LLM score-based evaluation less reliably distinguishes between right vs.\ wrong responses.\label{fig:evaluator}}
\end{figure*}

\subsection{Auto-Filter}
\label{sec:autofilter}

To estimate the quality of responses in the original dataset, 
CLEAR diverges from the conventional method of asking capable LLMs like ChatGPT to directly rate the input-output pair according to various criteria (e.g. \emph{helpfulness} as shown in Table \ref{tab:score_prompt}).
We instead employ LLM-derived confidence-estimates, specifically the BSDetector estimate introduced by \citet{DBLP:journals/corr/abs-2308-16175}. This estimates the confidence that a response is good in terms of two factors: \emph{observed consistency}  and \emph{self-reflection certainty}. 

BSDetector uses our same LLM to generate multiple candidate responses to a given prompt (via diversity-increasing techniques like temperature sampling and chain-of-thought), and then evaluates the semantic alignment between these candidate responses and the target response in the dataset (via natural language inference). Beyond this observed consistency, BSDetector  additionally integrates direct LLM self-evaluations of the target response (directly prompting the LLM to report its confidence that the response is good). The resulting confidence estimates account for both aleatoric and epistemic uncertainty, without requiring any modification/training of the LLM (no access to the LLM parameters is even required, enabling this approach to be used with arbitrary LLM APIs). Subsequent experiments reveal that this confidence-based approach to detect low-quality data is more precise than conventional LLM scoring of response quality (see Figure \ref{fig:evaluator}).

Given an instruction fine-tuning dataset of input-output pairs $\{(x_i, y_i)_{i=1}^{n}\}$, we use BSDetector with the base pretrained LLM (before it is fine-tuned) to estimate a confidence score $c_i$ for each pair $(x_i, y_i)$. We then filter out data pairs with low confidence scores below a predefined threshold $\gamma$: 
\[
F = \{(x_i, y_i) | c_i > \gamma \}.
\]
Subsequently, we fine-tune the LLM on the remaining training data $F$.

\subsection{Auto-Correct}
\label{sec:autocorrect}

Thus far, we considered filtering data estimated to be low-quality, but what if some of this data can be automatically rectified? A direct approach would be to substitute low-quality responses with LLM generated responses. For specialized domains, a pretrained general-purpose LLM like GPT-4 may be unable to generate  better responses for us to consider. But the LLM we fine-tuned after the Auto-Filter stage is specialized to our domain and should be able to generate some reasonable responses. If the auto-filtering was done well, then this fine-tuned LLM will exhibit less flaws being trained on less flawed data.

In the Auto-Correct stage, we proceed to generate responses  $\{(x_i, y_i')_{i=1}^n\}$ through queries to this fine-tuned model for each prompt $x_i$ in our dataset. What remains is to decide when the candidate response $y_i'$ generated by our current fine-tuned LLM is confidently better than the original dataset response $y_i$.
For this, we directly ask our base Foundation LLM (pre fine-tuning) via the LLM-as-judge prompt in Table \ref{tab:select_prompt}. As BSDetector is compatible any LLM, we can obtain confidence estimates for these LLM-as-judge preference predictions. For examples where the confidence (as estimated by BSDetector) that $y'_i$ is better than $y_i$ falls above a threshold $\eta$: we replace their target response with the LLM generated response and retain this pair in our curated dataset (rather than filtering it). This auto-corrected dataset is then used for further LLM fine-tuning, to yield a further improved model.

% \begin{figure}[tb!]
% \includegraphics[width=0.5\textwidth]{fig/select_prompt.pdf}
% \vspace{-1em}
% \caption{Prompt \citep{DBLP:journals/corr/abs-2306-05685} used to determine the preferable choice among $y$ and $y'$.\label{fig:select_prompt}}
% \end{figure}

\begin{table*}[ht]
\centering
{
\begin{tabular}{|p{14cm}|}
\hline
Please review two answers carefully and select the one that you believe is superior. Consider factors such as accuracy, completeness, relevance to the question. \\
Question: \texttt{[…]}  \\
You are provided with two responses to the same question: \\
\texttt{[The Start of Answer A]}: \texttt{[…]} \texttt{[The End of Answer A]} \\
\texttt{[The Start of Answer B]}: \texttt{[…]} \texttt{[The End of Answer B]} \\
Please provide a brief reasoning you used to derive it. After providing your explanation, output your final verdict by strictly following this format: “\texttt{[[A]]}” if Answer A is better, “\texttt{[[B]]}” if Answer B is better, and “\texttt{[[C]]}” for a tie.
\\
\hline
\end{tabular}}
\caption{Prompt \citep{DBLP:journals/corr/abs-2306-05685} used to determine the preferable choice among $y$ and $y'$.\label{tab:select_prompt}}
\end{table*}

\begin{table*}[ht!]
\centering
{\small
\begin{tabular}{@{\hspace{-1pt}}l@{\hspace{0pt}}l@{\hspace{2pt}}l@{\hspace{1pt}}clclclclclcl}
\hline
\hline
\multirow{3}{*}{} & \multirow{3}{*}{Training Data} & \multirow{3}{*}{Model} & \multicolumn{2}{c}{SQuAD-N} & \multicolumn{2}{c}{Email-N} & \multicolumn{2}{c}{DROP-N}\\
 &  &  & Valid JSON & Accuracy   & Valid JSON  & Accuracy  & Valid JSON  & Accuracy \\
 & or Prompting &  & (\%) &  (\%)  &  (\%) & (\%) & (\%) &  (\%) \\
 \hline
 \hline
\multirow{9}{*}{\parbox{2cm}{Pretrained \\ Model \\ (No  Fine- \\ Tuning)}} & \multirow{3}{*}{Zero-Shot} & GPT-3.5 & 99.85 & 66.65 & 93.5 & 23.25 & 99.50 & 33.40\\
 &  & GPT-4.0 & 99.90 & 75.93 & 100.0 & 48.25 & 100 & 39.80  \\
 &  & Llama-2 & 94.90 & 51.85 & 2.0 & 3.50 & 84.20 & 16.80\\
 \cline{2-9}
 & \multirow{3}{*}{One-Shot} & GPT-3.5 & 99.20 & 69.50 & 99.0 & 38.75 &99.60 & 40.80 \\
 &  & GPT-4.0 & 100.0 & 79.40 & 98.0 & 48.0 & 100.0 & 43.0 \\
 &  & Llama-2 & 24.65 & 9.70  & 17.25 & 19.50 & 32.0 & 4.90 \\
  \cline{2-9}
 & \multirow{3}{*}{Three-Shot} & GPT-3.5 & 87.60 & 61.20 & 95.75 &  47.0 & 98.50 & 41.80 \\
 &  & GPT-4.0 & 99.94 & \underline{80.08} & 100.0 & \underline{49.75} & 99.0 & \underline{46.10} \\
 &  & Llama-2 & 13.10 & 2.55 & 1.75 & 5.75 & 20.60 & 4.60 \\
\hline
\hline
\multirow{6}{*}{\parbox{2cm}{Fine- \\ Tuning}} & Original Data  & Llama-2 & 92.45 & 49.86 & 99.30 & 50.67 & 99.30 & 44.70\\
 & Auto-Filter Data & Llama-2 & 96.90  & 59.86 & 100.00 & 49.67 & 100.0 & 47.40 \\
  & Auto-Correct Data & Llama-2 & 96.90  & 71.44 & 99.67 &  52.33 &100.0 & 50.50\\
\cline{2-9}
 & Original Data  & GPT-3.5  & 97.90 & 64.50 & 100.0 & 43.0 & 100.0 & 56.80\\
 & Auto-Filter Data  & GPT-3.5   & 99.20 & 81.51 & 100.0 & 46.67 & 100.0 & 71.70 \\
 & Auto-Correct Data  & GPT-3.5  & 100.0 & \textbf{81.90} & 100.0 & \textbf{56.33} & 100.0 & \textbf{73.0} \\
 \hline
\hline
\end{tabular}}
\caption{Test set performance achieved by various Large Language Models when employed in non fine-tuning baselines or when fine-tuned. Both the model's ability to generate correct results (accuracy) and properly-formatted results (valid JSON \%) are reported. We underline the best non fine-tuning results, and indicate the best fine-tuning results in bold. Between each fine-tuning result, the training algorithm/code remains identical, only the underlying data is curated differently.}
\label{tab:1}
\end{table*}

\section{Experiments}

\begin{figure*}[ht]
\centering
\includegraphics[width=1.0\textwidth]{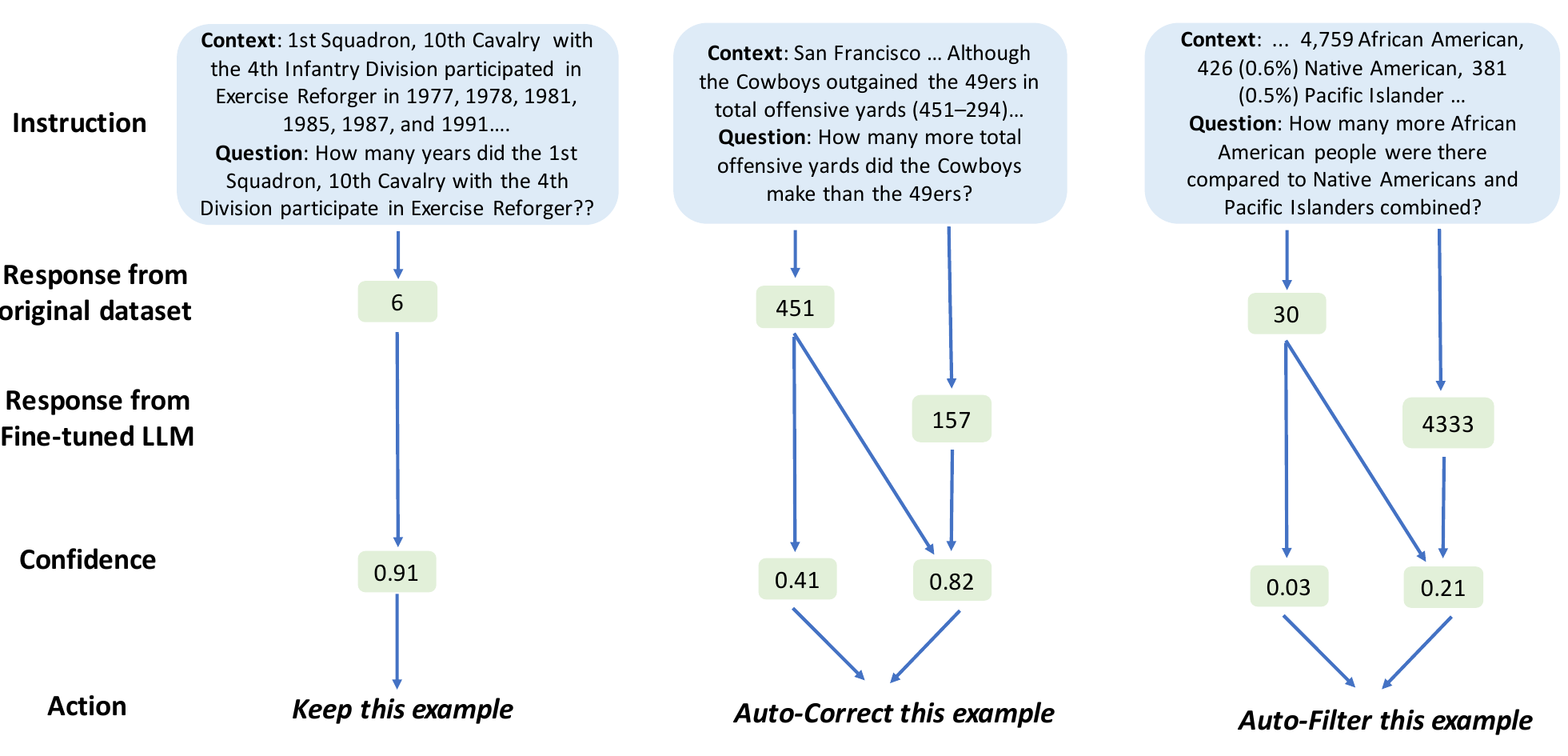}
\vspace{-1em}
\caption{Three examples from the \textbf{DROP-N} dataset. The first example (left) is retained in the dataset because the original response has high BSDetector-estimated confidence (0.91). 
The second example (middle) has an original response that is estimated to be low confidence (0.41), and the candidate alternative response generated from our fine-tuned LLM is better than the original response with confidence 0.82. Since this exceeds our confidence threshold $\eta=0.8$, we replace the target response for this second example with the LLM-generated candidate response in our curated dataset. 
The third example (right) has an original response that is estimated to be low confidence (0.03), but we also estimate low confidence (0.21) that the candidate  response from our fine-tuned LLM is better. This third example is thus entirely removed from our curated dataset. 
\label{fig:case_study}}
\end{figure*}

\paragraph{Datasets.} We evaluate the effectiveness of our data curation process across noisy versions of three supervised fine-tuning (text generation) datasets (see Figures \ref{fig:case_study},\ref{fig:case_study_2},\ref{fig:case_study_3}). \textbf{SQuAD-N} \citep{2016arXiv160605250R}: prompts are articles and target responses are answers to questions created by crowdworkers based on a collection of Wikipedia articles, with each answer being a specific text fragment or span from the related article. \textbf{Emails-N} \footnote{\url{https://huggingface.co/datasets/neelblabla/enron_labeled_emails_with_subjects-llama2-7b_finetuning}}: prompts are emails and target responses include categorizing the email into one of seven predefined themes by examining the email's subject and body content and also vary based on the email's length (whether the email content is short, medium, or long affects how the response is written). \textbf{DROP-N} \citep{DBLP:conf/naacl/DuaWDSS019}: prompts are articles and target responses are answers to reading comprehension questions that require  discrete reasoning over paragraphs (correctly answering requires resolving references in a question, perhaps to multiple places in the article, and performing basic operations over the references like addition, counting, or sorting). 

To study how our approach handles noisy data, we perturbed 20\% of each training dataset (not the corresponding test set). For the Emails dataset, the perturbation was to randomly swap target responses across different examples. To perturb a subset of the SQuAD and DROP datasets, where target responses are contained within a context passage in the provided instruction, we chose a random sentence from the context as the target response.

\paragraph{Evaluation metrics.} For each dataset, our LLM fine-tuning performance evaluation focuses on two metrics (computed over a fixed held out test set): how often the model's response format adheres to a valid JSON structure and how often the model's responses are correct. For each model produced via a fine-tuning method, we report the proportion of model responses that are in valid JSON format, and the accuracy of model responses (which is computed via the proportion of exact matches to target reference responses, since we expect a well-supervised model to able to match the types of target responses it was fine-tuned on).

\paragraph{Baseline Methods.} Our study also evaluates the following non fine-tuning methods: \textbf{Zero-shot} on GPT-3.5-turbo/GPT-4.0/Llama-2-7b-chat is directly querying these pretrained Foundation models. \textbf{Few-shot} on GPT-3.5-turbo/GPT-4.0/Llama-2-7b-chat is directly querying these pretrained Foundation models using in-context learning (with the indicated number of examples from the dataset inserted into each prompt as few-shot context). 
For the fine-tuning methods, we employ full model fine-tuning on Llama-2-7b-chat and OpenAI's GPT-3.5 Turbo fine-tuning API.
\textbf{Fine-tuning on the noisy data} refers to fine-tuning the model on the original datasets without any data curation. \textbf{Auto-Filter} refers to fine-tuning the model on a curated versions of the dataset, where data with low confidence levels have been eliminated as described in Sec.\ \ref{sec:autofilter}. This procedure sets the median confidence value across the dataset as the threshold $\gamma$, filtering out any data below this threshold. \textbf{Auto-Correct} refers to fine-tuning the model on curated versions of the dataset, where certain data has corrected responses generated as described in Sec.\ \ref{sec:autocorrect} (we set $\eta=0.8$). The fine-tuning routine stays the same when evaluating different data curation strategies -- we only alter the training dataset, not the model/  algorithm.

\paragraph{Other Details.} 
We study the effectiveness of data curation strategies across two different fine-tuning methods. On the Llama-2-7b-chat model, we conduct full model fine-tuning, in which all parameters of the neural network are updated via the Adam optimizer. We set the batch size at 128, and train for 3 epochs,  using a learning rate of $1 \times 10^{-5}$ with an accompanying cosine learning rate schedule. For the GPT-3.5 Turbo model, we use OpenAI's  fine-tuning API. The exact training algorithm/hyperparameters used remain undisclosed to us, but this API has been observed to be highly effective for LLM fine-tuning. 
When evaluating outputs from our models at test time, we perform all text generation with temperature 0, and limit the maximum number of output tokens to 512.

\section{Results}

Table \ref{tab:1} presents the results of our main experiments. Amongst the non fine-tuning approaches, GPT-4 stands out as the superior LLM, demonstrating the strongest performance across three datasets. For the pretrained GPT-4 model, few-shot learning outperforms zero-shot learning. But for the pretrained Llama-7B-chat model, few-shot learning produces much worse results compared to zero-shot learning, attributed to the smaller model's heightened sensitivity to the selection of few-shot demonstrations \citep{DBLP:conf/emnlp/ChenCZ023, DBLP:journals/corr/abs-2401-10529}. 

For the fine-tuned models, we observe that training on the entire noisy dataset without curation can even degrade model performance. Fine-tuning with only half of the data, refined through automatic filtering, yields better results than utilizing the complete, uncurated dataset. Moreover, training data curated via our Auto-Correct strategy further enhances model performance. Figures \ref{fig:case_study},\ref{fig:case_study_2},\ref{fig:case_study_3} depict for each dataset: a wrong response automatically identified in the Auto-Filter stage that was subsequently corrected in the Auto-Correct stage.

Our fine-tuned models can outperform even the most advanced model, GPT-4 with three-shot prompting. This highlights how even the most powerful LLMs may lack the capability to adequately address specific domain challenges. Unlike some other fine-tuning research, GPT-4 was not involved in any part of the data curation or training process underpinning our fine-tuned LLMs here.

\begin{table*}[ht]
\centering
{
\begin{tabular}{cclclclclcl}
\hline
\hline
 \multirow{3}{*}{Evaluator} & \multicolumn{2}{c}{SQuAD-N} & \multicolumn{2}{c}{Email-N} & \multicolumn{2}{c}{DROP-N}\\
 & Valid JSON & Accuracy   & Valid JSON  & Accuracy  & Valid JSON  & Accuracy \\
 & (\%) &  (\%)  &  (\%) & (\%) & (\%) &  (\%) \\
 \hline
 \hline
Random & 97.50 & 62.90 & 100.0 & 43.0 & 100.0 & 65.20\\
Score-based Evaluator & 99.50 & 78.40 & 100.0 & 39.67 & 100.0 & 73.00\\
Confidence-based Evaluator & 99.20 & 81.51 & 100.0 & 46.67 & 100.0 & 71.70  \\
 \hline
\hline
\end{tabular}}
\caption{Comparing different variants of the Auto-Filtering procedure. We try filtering the bottom 50\% of the data according to 3 different approaches: random scoring, score-based evaluator \citep{DBLP:journals/corr/abs-2308-06259}, and confidence-based evaluator. For each of the three resulting filtered dataset versions, we fine-tune the GPT-3.5 Turbo model and report its resulting performance. This experiment is repeated across SQuAD-N, Email-N, and DROP-N datasets.}
\label{tab:2}
\end{table*}

\begin{table*}[ht]
\centering
{
\begin{tabular}{cclclclclcl}
\hline
\hline
 \multirow{3}{*}{\parbox{4cm}{\centering Model used to generate the candidate response}} & \multicolumn{2}{c}{SQuAD-N} & \multicolumn{2}{c}{Email-N} & \multicolumn{2}{c}{DROP-N}\\
 & Valid JSON & Accuracy   & Valid JSON  & Accuracy  & Valid JSON  & Accuracy \\
 & (\%) &  (\%)  &  (\%) & (\%) & (\%) &  (\%) \\
 \hline
 \hline
GPT-3.5 Turbo & 99.20 & 77.80 & 100.0 & 6.0 & 100.0 & 63.00\\
Fine-tuned LLM & 100.0 & 81.90 & 100.0 & 56.33 & 100.0 & 73.0  \\
 \hline
\hline
\end{tabular}}
\caption{Comparing variants of the Auto-Correct procedure. We fine-tune a GPT-3.5 Turbo model 
on two datasets curated via Auto-Correct applied with candidate responses $y'$ generated from either: the pretrained GPT-3.5 Turbo base Foundation model, or the fine-tuned version of this LLM trained on our Auto-Filtered dataset. GPT-3.5 Turbo is also used as the model to estimate when candidate responses $y'$ are better than the original dataset responses $y$.}
\label{tab:3}
\end{table*}

\subsection{Estimating Response Quality in Auto-Filter}

We compare using a confidence-based response quality evaluator in our Auto-Filter procedure vs.\ an evaluator based on direct LLM scoring. The latter directly prompts the LLM (say GPT-3.5-turbo) to score a given input-output pair \citep{DBLP:journals/corr/abs-2308-06259} using a Likert scale rating from 1 to 5. Table \ref{tab:score_prompt} depicts the prompt used for this score based quality evaluation. After scoring the quality of each (instruction, response) pair in the dataset, we discard the 50\% with the lowest scores. Subsequently, we fine-tune the model on the remaining data. 

Table \ref{tab:2} presents results comparing this score-based approach against our confidence-based approach from Sec.\ \ref{sec:autofilter}. 
We additionally consider results based on fine-tuning the LLM on a randomly selected 50\% of the data.
Across all datasets, our confidence-based evaluator either matches or exceeds the performance of the score-based evaluator and random data selection, obtaining significantly better performance in the Email-N dataset.
% As discussed in \citet{DBLP:journals/corr/abs-2308-16175}, the confidence-based method involves directly asking the Large Language Model to assess whether a response is accurate or not. Additionally, it takes into account observed consistency, enabling it to provide a more precise evaluation of the response quality. 

\subsection{Using the LLM in Auto-Correct}

Here we consider a variant of our Auto-Correct stage, where we generate alternative candidate responses from the base pretrained Foundation model, instead of from our subsequently fine-tuned version of this LLM. Specifically we consider  GPT-3.5 Turbo to generate candidate responses $y'$ which are then fed into the same Auto-Correct procedure described in Sec.\ \ref{sec:autocorrect}. Table \ref{tab:3} reveals that using the fine-tuned version of this LLM to generate candidate responses performs better across all datasets.

% In the Auto-Correct process, we choose the preferred response by comparing the original response with that generated by fine-tuned LLMs. Consider an alternative approach where we obtain responses from a different, advanced LLM, such as ChatGPT, and then choose the more preferable response between the original and ChatGPT's. Specifically, we use GPT-3.5 Turbo to generate responses to all prompts, and then select the preferable one using the prompt in Table \ref{tab:select_prompt}. The findings, shown in Table \ref{tab:3}, illustrate that our method still exceeds the efficacy of choosing responses from GPT-3.5 Turbo. This emphasizes the importance of employing fine-tuned LLMs to produce candidate responses in the Auto-Correct process. As demonstrated in Table \ref{tab:1}, the inferior performance in a zero-shot setting already suggests that relying on other LLMs for generating candidate responses might not be a viable strategy. Furthermore, considering that we use a model from the GPT family to select the preferable response, there could be an inherent bias towards responses it generates itself. 

\section{Conclusion}

This paper presents a general pipeline for curating better versions of an existing instruction fine-tuning dataset. Our data-centric CLEAR approach can be combined with any model and fine-tuning algorithm. While better models and fine-tuning algorithms will inevitably be invented, data-centric approaches like ours can remain useful. As future LLMs advance, their ability to curate the data via CLEAR will advance, facilitating \emph{even better} LLMs to be trained on this better curated data.

Experiments demonstrated that our data curation process produces substantial improvements in the performance of fine-tuned LLMs across different noisy datasets, models, and training algorithms (without being tailored to each setting). 
While our approach fixes issues in an existing dataset, augmenting this data with additional \emph{synthetic} examples is another data-centric approach that appears promising to combine with CLEAR.

% Our data curation process produces substantial improvements in the performance of fine-tuned LLMs across different datasets, models, and training algorithms. One open question is how can automated data curation be effectively integrated with other data augmentation techniques to further enhance model performance? For instance, creating additional synthetic datasets in situations where available data is scarce, and how to integrate these synthetic datasets with the original dataset to further enhance the performance of the fine-tuned model.

\section*{Limitations}
While our automated data curation pipeline presents a significant advancement in enhancing the quality of instruction tuning datasets for large language models (LLMs), it is important to acknowledge its limitations. The pipeline's current framework does not explicitly account for the possibility of biases within the original dataset or those introduced during the automated curation process. Since the model's performance and the quality of its output are contingent upon the data it was trained on, any inherent biases could be perpetuated or amplified through successive iterations of fine-tuning and correction.

% \clearpage

\bibliography{anthology,custom}
\bibliographystyle{acl_natbib}

\clearpage

\appendix
\onecolumn

\section{Prompt for Score-based Answer Quality Evaluator}

\begin{table*}[ht]
\centering
{
\begin{tabular}{|p{14cm}|}
\hline
 Below is an instruction from an user and a candidate answer. Evaluate whether or not the answer is a good example of how AI Assistant should respond to the user’s instruction. Please assign a score using the following 5-point scale: 1: It means the answer is incomplete, vague, off-topic, controversial, or not exactly what the user asked for. For example, some content seems missing, numbered list does not start from the beginning, the opening sentence repeats user’s question. Or the response is from another person’s perspective with their personal experience (e.g. taken from blog posts), or looks like an answer from a forum. Or it contains promotional text, navigation text, or other irrelevant information. 2: It means the answer addresses most of the asks from the user. It does not directly address the user’s question. For example, it only provides a high-level methodology instead of the exact solution to user’s question. 3: It means the answer is helpful but not written by an AI Assistant. It addresses all the basic asks from the user. It is complete and self contained with the drawback that the response is not written from an AI assistant’s perspective, but from other people’s perspective. The content looks like an excerpt from a blog post, web page, or web search results. For example, it contains personal experience or opinion, mentions comments section, or share on social media, etc. 4: It means the answer is written from an AI assistant’s perspective with a clear focus of addressing the instruction. It provide a complete, clear, and comprehensive response to user’s question or instruction without missing or irrelevant information. It is well organized, self-contained, and written in a helpful tone. It has minor room for improvement, e.g. more concise and focused. 5: It means it is a perfect answer from an AI Assistant. It has a clear focus on being a helpful AI Assistant, where the response looks like intentionally written to address the user’s question or instruction without any irrelevant sentences. The answer provides high quality content, demonstrating expert knowledge in the area, is very well written, logical, easy-to-follow, engaging and insightful. Please first provide a brief reasoning you used to derive the rating score, and then write "Score: " in the last line. \\ Input: [] \\ Response: []
\\
\hline
\end{tabular}}
\caption{Prompt that \citet{DBLP:journals/corr/abs-2308-06259} use to have a LLM to directly score instruction-response pairs.}
\label{tab:score_prompt}
\end{table*}

\begin{figure*}[ht]
\centering
\includegraphics[width=1.0\textwidth]{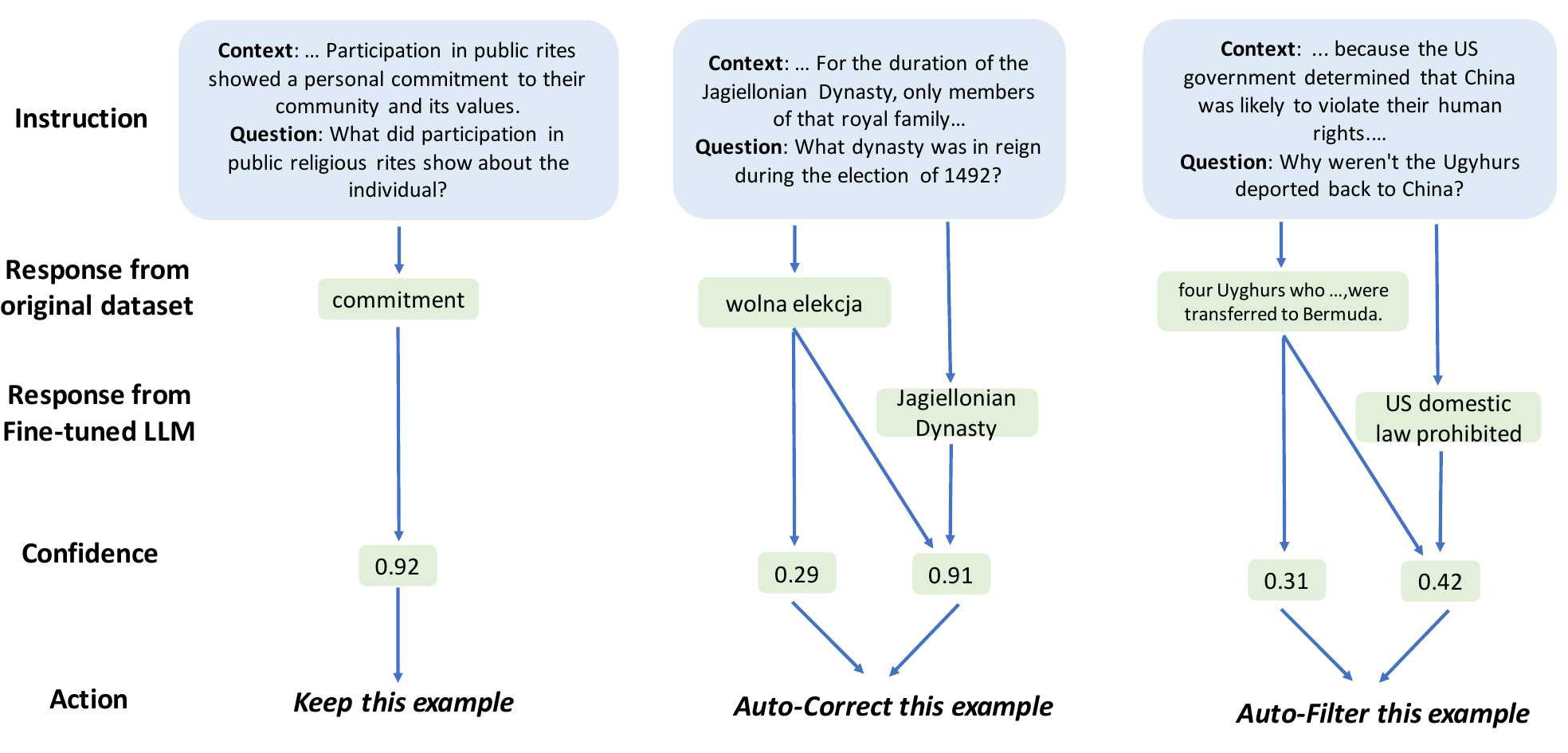}
\vspace{-1em}
\caption{Three examples from the \textbf{SQuAD-N} dataset. The first example (left) is retained in the dataset because the original response has high BSDetector-estimated confidence (0.92). 
The second example (middle) has an original response that is estimated to be low confidence (0.29), and the candidate alternative response generated from our fine-tuned LLM is better than the original response with confidence 0.91. Since this exceeds our confidence threshold $\eta=0.8$, we replace the target response for this second example with the LLM-generated candidate response in our curated dataset. 
The third example (right) has an original response that is estimated to be low confidence (0.31), but we also estimate low confidence (0.42) that the candidate  response from our fine-tuned LLM is better. This third example is thus entirely removed from our curated dataset.\label{fig:case_study_2}}
\end{figure*}

\begin{figure*}[ht]
\centering
\includegraphics[width=1.0\textwidth]{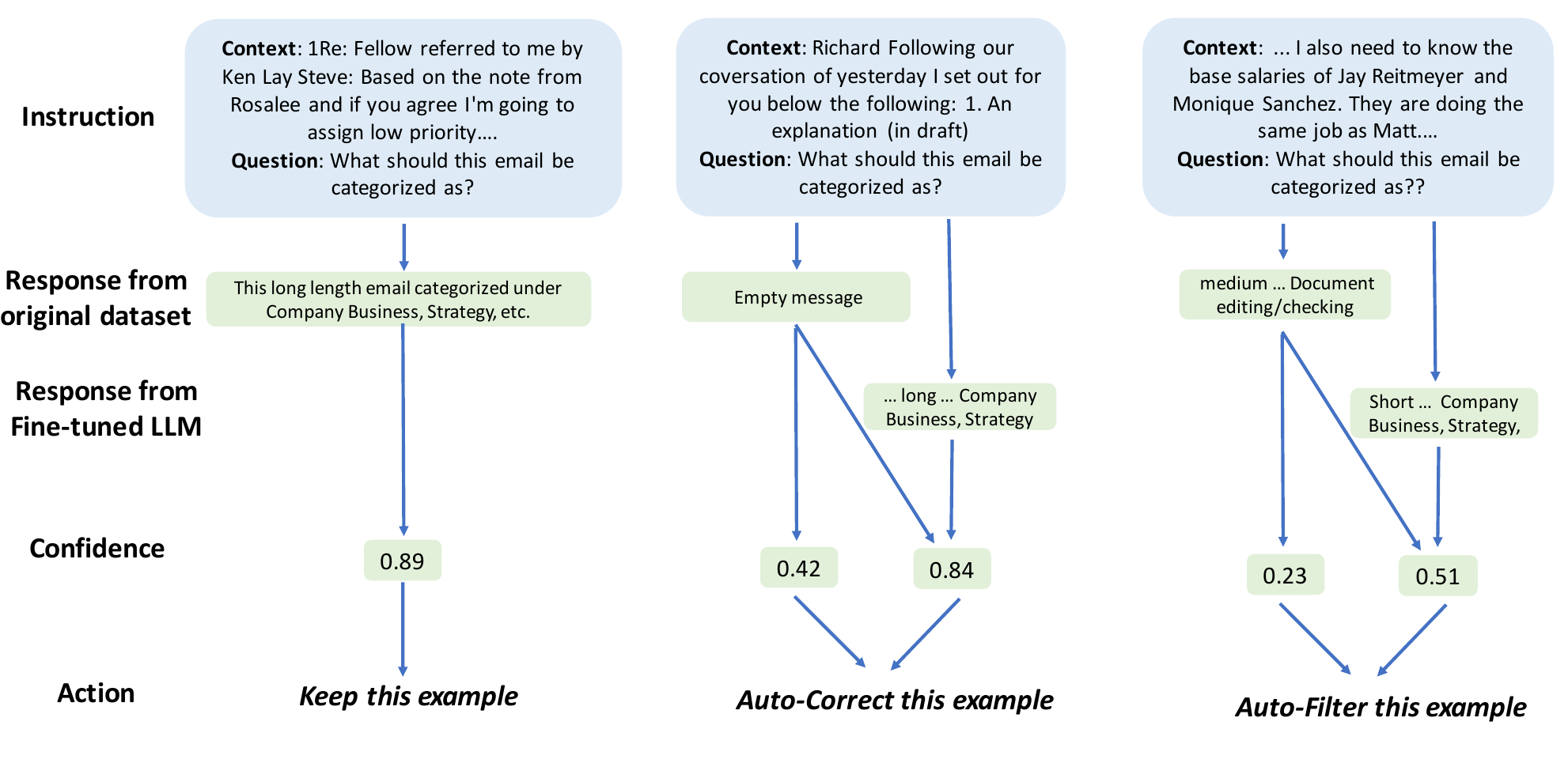}
\vspace{-1em}
\caption{
Three examples from the \textbf{Email-N} dataset. The first example (left) is retained in the dataset because the original response has high BSDetector-estimated confidence (0.89). 
The second example (middle) has an original response that is estimated to be low confidence (0.42), and the candidate alternative response generated from our fine-tuned LLM is better than the original response with confidence 0.84. Since this exceeds our confidence threshold $\eta=0.8$, we replace the target response for this second example with the LLM-generated candidate response in our curated dataset. 
The third example (right) has an original response that is estimated to be low confidence (0.23), but we also estimate low confidence (0.51) that the candidate  response from our fine-tuned LLM is better. This third example is thus entirely removed from our curated dataset.\label{fig:case_study_3}}
\end{figure*}

% \section{Example Appendix}
% \label{sec:appendix}

% This is a section in the appendix.

\end{document}